# A Comparative Study of Machine Learning Methods for Verbal Autopsy Text Classification

Samuel Danso[1], Eric Atwell[2] and Owen Johnson[3]

[1]Language Research Group, School of Computing, University of Leeds
LS2 9JT, U.K.
scsod@leeds.ac.uk

[2]Language Research Group, School of Computing, University of Leeds
LS2 9JT, U.K.
E.S.Atwell@leeds.ac.uk

[3]Yorkshire Centre for Health Informatics, University of Leeds
LS2 9JT, U.K.
O.A.Johnson@leeds.ac.uk

**Abstract**

A Verbal Autopsy is the record of an interview about the circumstances of an uncertified death. In developing countries, if a death occurs away from health facilities, a field-worker interviews a relative of the deceased about the circumstances of the death; this Verbal Autopsy can be reviewed off-site. We report on a comparative study of the processes involved in Text Classification applied to classifying Cause of Death: feature value representation; machine learning classification algorithms; and feature reduction strategies in order to identify the suitable approaches applicable to the classification of Verbal Autopsy text. We demonstrate that normalised term frequency and the standard TFiDF achieve comparable performance across a number of classifiers. The results also show Support Vector Machine is superior to other classification algorithms employed in this research. Finally, we demonstrate the effectiveness of employing a 'locally-semi-supervised' feature reduction strategy in order to increase performance accuracy.

## 1.0 Introduction

Text Classification (TC) is an automated process of assigning textual documents to a set of predefined categories. This process has seen unprecedented growth in interest and research due to the abundance of documents available in textual format. The process is cross-disciplinary as it encompasses several subfields under the umbrella of computer science: Natural Language Processing (NLP), Machine Learning, Pattern Recognition, and Statistical theories[1]. There is a continued effort by the research community with the aim of improving the classification accuracy of machine learning classification algorithms by exploring the various subfields. This is due to the fact that numerous factors determine the performance of a given classifier, and these include: the data and domain; machine learning algorithm; and the features and their representation schemes employed in the process of building a classifier for the classification task[2].

The biomedical domain is one area that is witnessing a high rate of growth in research in the application of TC technology [3-5]. However, TC research has not been extended to Verbal Autopsy (VA) narratives, which are considered another subtype of biomedical genre [6] . VA is an alternative approach to determining Cause of Death(CoD). It is a World Health Organisation (WHO) recommended approach being applied in developing countries where the majority of deaths occur outside health facilities[7]. Ideally, a Cause of Death should be certified by a doctor, but there are insufficient medical staff to administer autopsies in all such cases. Instead, a non-clinician field-worker goes to interview a close relative of the deceased about the circumstances of the death. The Verbal Autopsy is the written record of this interview. Currently, the VA must be assessed by clinicians off-site, who determine likely CoD by reviewing the interview record. A method for automatic classification of VAs according to CoD offers numerous potential benefits: relatively lower information [8, 9]. It has however been established that the coded part is invariably limited in capturing all the available information; thus, the need for free text as an alternative approach[7]. This paper reports on automatic approaches carried out that focus on the free text part of the VA information.

Danso et al [6] discuss the possible challenges associated with using machine learning approaches to classification of Verbal Autopsy text. In brief, VA is a nonstandard text generated from dialogues between two non–medically trained people. Consequently, the text is characterised with issues that may not be found in a standard biomedical text which include: non-standard medical terms; non-medical expressions; spelling and grammatical issues; and use of local terms to describe medical conditions. How to effectively deal with text of this nature in order to achieve good classification accuracy remains a challenge in medical informatics.

This paper carries out a comparative study on the various aspects of TC processes in order to identify suitable approaches for the classification of Verbal Autopsy text. The paper investigates various feature value representation schemes, machine learning algorithms, and the effect of feature reduction on the overall performance accuracy of a machine learning algorithm. To the best of our knowledge this is first paper that reports on a comparison study on machine learning approaches to classify Verbal Autopsy text.

### 1.1 Feature Value representations

Feature value representation, which is also referred to as Term Weighting is a transformation technique that allows documents to be directly interpreted by machine learning classifiers. It is a technique proposed by Salton et al [10] to represent documents as a feature vector, popularly employed in information retrieval and now being applied to TC. A feature of a document could either be a word, phrase or in any other form used to identify the content of the document. Regardless of the scheme of representation, each feature must be associated with a value or weight, which indicates the importance of the feature in terms of its contribution to the classification. As argued by [11] the weighting strategy employed has major implications for the accuracy of classification than the choice of learning algorithm employed in the classification process. There are numerous term weighting schemes proposed in the literature by various researchers [12-14] However, all these weighting schemes are variants of the three basic and standard schemes as summarised below:

Table 1: Term weighting schemes

| Scheme | Description |
|--------|-------------|
| **Binary** | Boolean logic representation; 1 = present, 0 = not present |
| **TF** | Frequency count of terms found in a given document |
| **DF** | Frequency count of documents that contain a given term. |

With the exception of the binary approach, which represents feature occurrence as '1' and non-occurrence as '0', the other two approaches suggest weights based on frequency counts of either the feature or the documents containing the feature. The basic assumptions here are that the importance of a feature is based on its frequency of occurrence in a given document (TF), and a count of documents of which that feature occurs (DF). While these schemes are sometimes employed as stand-alone, they are also sometimes mathematically combined. For example the DF and TF are mostly combined by the product of the TF and the inverse of DF (iDF) to form another widely used scheme known as TFiDF[10]. The idea for this combination is that the higher the frequency of a term in a given document, the more it is a representative of its content. Also, the more documents a term occurs in, the less powerful it is in discriminating between a given set of documents [15]. Recent advancement in research in this subfield has seen more sophisticated approaches; a combination of feature selections metrics such as information gain, chi-square, gain ratio and odd ratios with TF and DF have been explored[12, 13]. This has led to categorisation of term weighting schemes into supervised and unsupervised methods due to the process employed in estimating the values [14]. Furthermore, DF or TF are sometimes combined with a normalisation factor. For example a normalised factor of document length takes into account terms of the same frequency in different documents to ensure features found in both short and long documents are of equal importance[16]. The investigations carried out in this paper considered the standard term weighting schemes: Binary; Term Frequency; the standard TFiDF; and Normalised Term Frequency, which is normalised by the length of the VA document due to the varying length of the Verbal Autopsy documents.[6]

**1.2 Machine Learning Classification techniques**

The selection and creation of a machine learning classifier is the next step once the document representation scheme is finalised. Numerous machine learning techniques have been employed to tackle various classification problems[17]. One of the main differences that exist between these techniques is the philosophy behind the learning process. We discuss some of the machine learning techniques that have successfully been employed in TC in which we investigate their performance in our experiments: Naïve Bayes; Support Vector Machines; and Decision Trees.

Naïve Bayes (NB) is considered to be a relatively simple machine learning technique based on probability models- Bayesian theorem[18]. This classification technique analyses the relationship between each feature and the class for each instance to derive a conditional probability for the relationships between the feature values and the Class. The conceptual framework for NB is based on joint probabilities of features and Classes to estimate the probabilities of a given document belonging to a given Class.

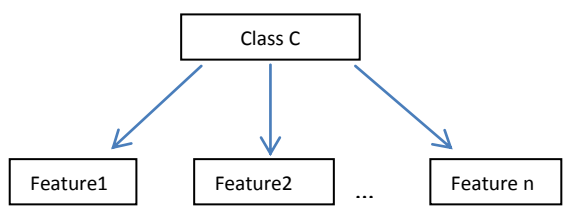

Figure 1: Naïve Bayes conceptual representation

During training, the probability of each Class is computed by counting how many times it occurs in the training dataset known as the "prior probability". In addition to the prior probability, the algorithm also computes the probability for the instance 'x' given a Class 'c' with the assumption that the features are independent. It is considered naïve due to the assumptions that is holds about the independence of conditional probabilities of words found in a given document of a given Class[19].

This probability becomes the product of the probabilities of each single feature. The probabilities can then be estimated from the frequencies of the instances in the training set. Numeric attributes can have a large number (possibly infinite) of values and the probability cannot be estimated from the frequency distribution, which tends to reduce the performance of Naïve Bayes. However NB has proved to be robust to noise and missing data as it has the ability of performing the probabilities without having any impact on the final outcome[20]. Its relative simplicity is also an indication of why it tends to be more popular than the majority of the classification techniques found in the literature[21].

Support Vector Machine (SVM): this classification technique is relatively the newest among the supervised machine learning techniques found in the literature[17]. SVM has proven to be robust in dealing with noisy and sparse datasets, and as result, has been a preferred choice to be employed in solving various classification problems. SVM was originally proposed by Vapnik in 1999 to deal with classification problems, and the principles under which SVM operates could be described as a hybrid of linear and non-linear, which is based on the Structural Risk Minimisation principle[22]

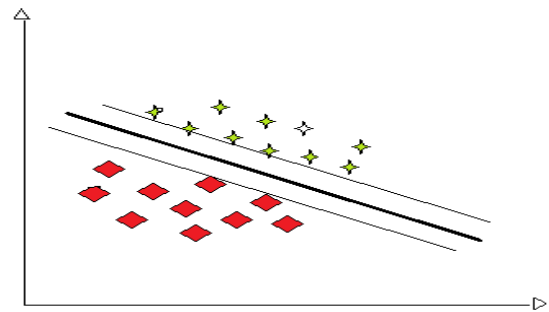

Figure 2: graphical representation of SVM learning algorithm

During learning, SVM employs a technique of 'maximal-margin-hyper-plane', where the maximum linear distance between Classes in the features space is estimated and separated from each other. However, where this cannot be achieved because non-linearity exists, SVM has the ability to adapt by employing 'kernels' that are able to map the non-linearity between Classes or categories and feature space. The resulting hyper-plane established in the feature space by this kernel provides a direct mapping to non-linear structure that exists within the feature space[17]. Despite its powers discussed above, SVM tends to be computationally expensive by virtue of the kernel technique it employs during learning. This however can be minimized during SVM model training and evaluation since the kernel is a parameter that can be adjusted depending on the performance, which eventually reduces computational cost.

Decision Tree (DT): DT has been employed successfully in many traditional applications in different domains [23]. Despite the fact that it can be regarded as relatively old technique, DT has stood the test of time. For example, DT has recently been employed as a machine learning technique to develop classification models that automatically classify pancreatic cancer data[24]. DT based algorithm 'learns' from training examples by classifying instances and sorting them based on feature values. Each node in a DT represents a feature of an instance to be classified, and each branch represents a value that the node can include in making a decision. The figure below is an illustration of how DT works within the feature space.

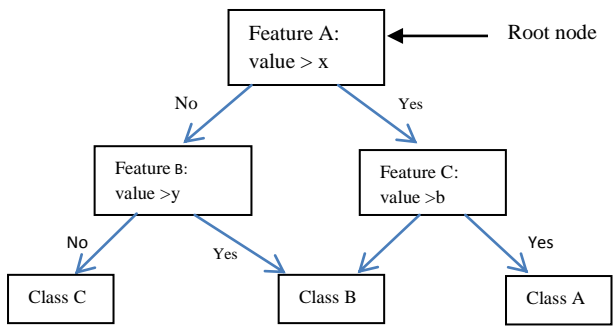

Figure 3: graphical representation of DT learning algorithm

The algorithm starts the process at a root node of the tree. This root node is established by finding the feature that best divides the feature space, and there are numerous approaches to identifying the best feature[17]. The Classes are assigned based on weights that are computed on the features during the processes of learning and these weights are used to classify unseen data. Due to the approach DT uses to search for a solution within the problem space, efficiency tends to be an issue, especially when dealing with large datasets. This has resulted in research into how this could be improved. Nevertheless, DT is characterised by its relative transparent outputs, which are easy to be read and understood by humans. DT has been shown to have superior performance over other techniques with regard to some specific domains with datasets that have discrete/categorical data type attributes[25].

### 1.3 Feature Reduction techniques

Feature reduction is a major activity in the TC process, as it seeks to reduce the high dimensionality of feature vectors that mostly results in high computational cost and adversely affecting the performance of learning algorithms. For example it is suggested that number of features should not exceed number of training examples as this nature may trigger over-fitting to occur[15]. Moreover, it has subsequently been well established that a strategic removal of irrelevant and redundant features tends to increase efficiency and performance accuracy of a machine learning algorithm [26]. Consequently, this has led to the integration of feature reduction as part of the steps for many machine learning algorithms[12]. Decision tree is an example of a learning algorithm that identifies "important" features to serve as nodes that discriminate between categories.

Lui et al[12] groups the feature reduction approaches into two: global and local. The global approach allows features to be identified that are discriminative across all categories. The local approach on the other hand allows features that are indicative of each category to be selected. Various works have explored both global and local based approaches to feature reduction with the aim of identifying the best amongst them. Evidence from the literature demonstrates their relative advantages and disadvantages and their performance tends to largely depend on the dataset [12, 27, 28]. We explore the local approach to feature reduction to investigate the effect of this approach on the classification of VA text. Our approach, referred to as 'locally-semi-supervised' is however different from the local based approaches reported in the literature. We employ a log-likelihood statistical metric, which is a variation of the different metrics employed so far in the literature to identify the possible features that are indicative of each CoD category in this dataset. A detailed description of the method is given in the subsequent section of the paper.

### 2.0 Methods
### 2.1 Dataset

The experiment involves a total of 6407 Verbal Autopsy documents, consisting of two levels of groupings: the higher level has 5 categories and the fine grained level consists of 16 CoD categories. See Danso et al [6] for a detailed description of the dataset. This experiment however focuses on the higher level of groupings. The table 2 below is a breakdown.

Table 2: statistics of dataset

| Categories | Number of documents | % distribution |
|---|---|---|
| Neonatal | 2005 | 31.3 |
| Non_stillbirth_unknown_cause | 801 | 12.5 |
| Intrapartum_still_birth | 998 | 15.6 |
| Antepartum_stillbirth | 1376 | 21.5 |
| PostNeonatal | 1227 | 19.1 |
| Total | 6407 | 100 |

### 2.2 Pre-possessing and experimental setup.

The text was converted to lower case and tokenised by whitespaces. All punctuations were also removed. Even though stop-words are removed during the pre-processing stage in most NLP tasks under the pretext that they are not informative and subsequently non discriminative, this however has led to mixed and inconclusive results[29]. Also [27] argues that stop-words tend to be domain specific, so the stop-words were therefore not removed from the dataset for this experiment.

Separate datasets were prepared based on feature value representations under investigation: Binary, Term Frequency; Normalised Term Frequency, expressed as the Term Frequency divided by the total number of Terms found in the given document(document length); and TFiDF. The files were stored in a format readable by the WEKA Machine Learning software[20] used in carrying out this experiment. WEKA has implementations of the machine learning algorithms discussed above, and thus employed in carrying out the experiments were: the Naïve Bayes algorithm developed by [30]; the Platt's Sequential Minimal Optimisation(SMO), which is a variant of the standard SVM algorithm[31]; and the Random Forest, a variant of the Standard Decision Tree algorithm [32]. All learning parameter default values set by WEKA for these algorithms were not changed.

Our 'locally-semi-supervised' feature reduction techniques employed the log-likelihood estimation metric in the feature reduction process due to its superiority over the other metrics as pointed by [33]. To achieve this training set the corpus was split into the five CoD categories. Each sub dataset was compared against the whole based on their log-likelihood ratio using the AntConc software[25] . This enabled us to rank all words that are indicative of a given CoD category. This process was repeated in turn for all five categories. Various thresholds levels (top 10, 25, 50, 100, 150, 200, 250, 300, 350, and all) words were selected based on the rankings generated for each category and combined for the experiment.

### 2.3 Evaluation metrics

Precision, Recall and F1 score are employed as the standard metrics to evaluate the performance machine learning methods. However, two types of measurements exist for F1-score: Micro – averaging and Macro-averaging. The former is used when there is an even distribution of Classes in the dataset. We employ Macro – averaging to determine the overall performance due to the highly uneven distribution of the multi-class dataset being used; it allows equal weights to be computed for each CoD category[34].

### 3. Results and Discussion

The experiments employed the 10 fold cross validation evaluation method to allow a random split stratified by the categories into training and test sets for 10 runs[35]. A weighted average is then computed over the 10 folds as shown in table 1.

Table 1: Macro-F1 average score results obtained from different machine learning classification algorithms on various feature value representation schemes.

| Feature Value Representation | Random Forest | Naive Bayes | Support Vector Machine |
|---|---|---|---|
| Binary | 0.149 | 0.255 | 0.149 |
| Frequency | 0.149 | 0.363 | 0.391 |
| Normalised Frequency | 0.149 | 0.39 | 0.416 |
| TFiDF | 0.149 | 0.373 | 0.419 |

### 3.1 Feature Value representation

The results in Table 1 demonstrate the variations that exist in performance between algorithms and feature value representations. As seen, Random Forest achieved the worst performance whereas the SVM achieved the best performance, which is followed by Naïve Bayes across all the feature value representation schemes under study. With regards to feature value representation schemes, the Binary scheme achieved the worst performance across all the three learning algorithms. This is followed by the Term Frequency scheme. Normalised Term Frequency and TFiDF however achieved comparable performance with SVM. In contrast, Normalised Frequency achieved about 2 % higher over TFiDF for Naïve Bayes algorithm. Also, as seen, Naïve Bayes achieved better performance over SVM under the Binary representation scheme.

The evidence from the results tends to demonstrate that the choice of feature value representation has implications for the performance of learning algorithms. The poor performance of the binary representation scheme suggests that it is not an appropriate scheme for VA text. The possible reason that may account for this is the uncontrolled vocabulary nature of the of VA text resulting in terms being rare and resulting in a sparse dataset. Binary scheme may be an appropriate scheme for dataset with controlled vocabulary where limited variations in concept may exist. For example Gamon[36] chose binary feature representation and yet achieved good results because of the brevity of the documents (which may result in a small number of unique words or limited vocabulary) used in carrying out the experiments. A similar reason could be attributed to the Term Frequency scheme. However because there appear to be some weights in terms of the frequency count of how each term appears in the document, this extra information was useful for the learning algorithm, which resulted in an improvement over the binary scheme.

The comparable results obtained between Normalised Term Frequency and TFiDF was quite surprising considering the fact that the IDF normalisation factor tends to assign lower values to common terms that occur in several documents[37] such as the stopwords, and consequently resulting in a better performance accuracy[10], but these words were not removed from the dataset. This suggests that there were relatively limited occurrence of the so-called common words due to misspellings; thus, resulting in variations of the same word and consequently rare but equally important in discriminating, and therefore re-enforcing the statement "Little words can make a big difference for text classification" by Rillof [29]. This result is a confirmation of our initial exploratory experiments, which suggested that removal of stopwords has adverse impact on the performance of classifier. For example the term "during", which is considered a stop-word in English, but appears to be a keyword that describes delivery events which distinguishes between intra-partum and antepartum stillborn. However, the computational cost associated with the generation of TFIDF values tend to be considerably higher than the Normalised Frequency. This suggests Normalised Frequency as the suitable scheme for the VA text classification.

### 3.2 Machine Learning classification algorithms

Although Random Forest has successfully been applied to classify the coded part of Verbal Autopsy data[38], the results obtained from this experiment suggest that it is not an appropriate choice for classification of VA free text. The differences that exist between the feature vectors generated from the coded and the free text data may account for this. The coded data feature vector is derived from a controlled vocabulary with a limited number of features; possibly a list of questions with 'Yes' and 'No' answer options. In contrast, the uncontrolled vocabulary characteristic of VA free text results in a large number of features. Since Decision Trees have generally been found to be susceptible to over-fitting with 500 or more features [39], there is the possibility that this may have harmed the Random Forest algorithm since the text tends to generate a high number of features. The closed part of VAs is unlikely to exceed the 500 features limit and may therefore be suitable for the Random Forest learning algorithm.

The independent assumption applied in Naïve Bayes may explain the relatively better performance compared with Random Forest, and even performing better than SVM for the binary representation scheme. This is because Naïve Bayes calculates the probability of a document belonging to a class by multiplying the probability of all the feature values, for both word occurrences ('1') and non-occurrence('0') in the document, and coupled with the highly biased characteristic, Naïve Bayes tends to be susceptible to skewed data, which results in it achieving a relatively better overall accuracy for skewed data[40]. The '1' values of the binary representation for the majority Class (Neonatal) may have outweighed the other 4 Classes in this case, resulting in a better score than SVM.

However, with the exception of the binary representation, the consistent superior performance of the SVM algorithm over both Naïve Bayes and Random Forest algorithms is not surprising. SVM has been consistently shown to have relatively better performance in Text Classification experiments[20], and the results from this experiments are not an exception. This outstanding performance of SVM could be attributed to a number factors: majority of TC problems are mostly linearly separable, and SVM employs threshold functions to develop margins that linearly separate the Classes; SVMs tend to use over-fitting protection mechanism that is independent of the dimensionality of the feature space, thus, the number of features tends not to be an issue; and SVMs are well designed to deal with sparseness found in feature vectors[39]. Danso et all's [6] description of the VA text seems to correlate with the taxonomy of issues outlined that the SVM algorithm is designed to address. It therefore seems natural that SVM tends to perform better than Naïve Bayes and Random Forest for this task.

### 3.3 Feature reduction

Having identified SVM as the best performing algorithm for this domain, the feature reduction experiment considered only SVM. Figure 4 below therefore shows results obtained from SVM when experimenting over a number of feature reduction thresholds.

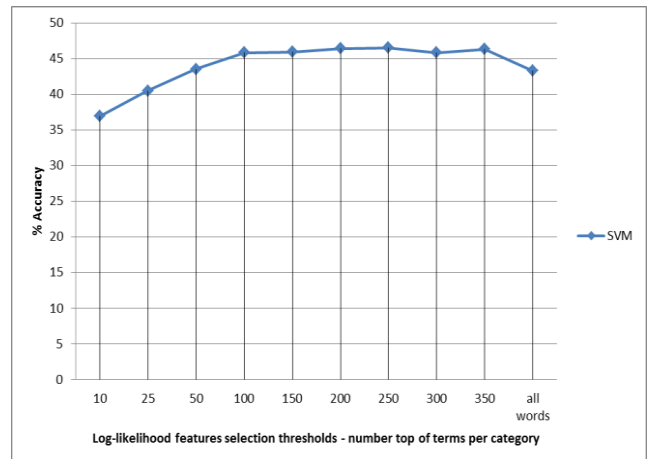

Figure 4: results on various feature reduction thresholds

Figure 4 shows performance accuracies obtained from various thresholds beginning with the top 10 words of each of the 5 categories as per the log-likelihood rankings. As seen, performance accuracy increased as the number of features increased. However, there was a change in trend as the rate of increase flattened between 100 and 300 top features, with the top 250 features achieving the highest of a marginal increase of 0.1%. This trend begins to decrease from the 300 top features achieving about 3.6 % less than the top 250 features when all the features were considered.

SVM robustness to over-fitting has resulted in the argument within the research community that it is irrelevant to carry out feature reduction before training[27]. The results from the feature reduction suggest that there are additional benefits to reducing features as a prior step to performing learning. The substantial increase in performance accuracy may be due to the removal of noisy features; and additional information presented to the learning algorithm as a result of the feature reduction process employed in this experiment. The 'locally-semi-supervised' approach employed may have effectively selected features (words) that have stronger correlation with the CoD categories. The result has demonstrated that an appropriate feature reduction strategy may improve the performance of the SVM, which is similar to other feature reduction experiments[40].

### 4.0 Conclusion and future work

This paper has presented results of a comparative study carried out to explore three aspects of machine learning approaches suitable for the classification of Verbal Autopsy text: feature value representation; machine learning algorithms; and the effect of feature reduction. The experimental results suggest that Normalised Term Frequency performance is comparable to the standard TFiDF, but Normalised Frequency may be the best option when computational cost of generating TFiDF values is taken into account. Binary and Term Frequency were also explored but were found not to be suitable. The SVM algorithm was found to be the best performing algorithm and the most suitable for VA text. However, Naïve Bayes was found to outperform SVM and Random Forest when explored with binary feature representation, which may be appropriate for data with limited vocabulary size such as the VA closed part. The experiment also shows that employing a 'locally – semi-supervised' approach to reducing features resulted in a substantial improvement in accuracy.

Although researchers have explored the closed part of the VA data, we have not attempted to compare results reported from those works with the results obtained from this experiment. This is because the focus of this paper was to establish the best obtainable baseline results from the methods explored using a Bag-of-Words approach, which will serve as a building block to constructing a classifier with higher accuracy using machine learning approaches. Thus, future work will explore the feature space to identity features that will lead to an improved accuracy of the SVM algorithm. Additionally, the method employed in carrying out the feature reduction seems to have performed as expected. However, it may be good to employ other feature reduction approaches to compare with the approach employed in this experiment and therefore future work could explore this possibility.

**Acknowledgments**
Our special thanks goes to Professor Betty Kirkwood of the London School of Hygiene and Tropical Medicine and entire trial management team of the ObaapaVita and Newhints projects that generated the dataset used in carrying out this study. Our appreciation also goes to the Commonwealth Scholarship Commission for their funding support for this research.

**First Author.** Samuel Danso is currently pursuing a PhD at the University of Leeds in the United Kingdom. He holds a BSc(Hons) in Computing and an MSc in Advanced Software Engineering. He has over 10 years experience in database design and implementation for large and complex epidemiloigical and clinical studies carried out by the London School of Hygiene and Tropical Medicine in colloboration with the Kintampo Health Reseach Centre, Ghana, where the studies are conducted. Samuel is a Commonwealth Scholar, and has co-authored a nmber of peer-reveiwed journal publications. His research interest lies in Text Analytics, particularly focused on Health Informatics.

**Second Author:** Eric Atwell is an Associate Professor at the University of Leeds in the United Kingdom. He has over 30 years experience in conducting and supervising language research projects. His research specialty is in the area of Corpus Linguistics and Text Analytics: Machine Learning and Data Mining analysis of a corpus of text - in English, Arabic, or other languages - to analyse the text and detect "interesting" and "useful" features or patterns. He has led research projects supported by various funding bodies including the EPSRC, ESRC, CPNI, HEFCE, MoD, and industry.

**Third Author:** Owen Johnson is Senior Fellow at the University of Leeds in the United Kingdom. He has over 20 years experience as a practitioner and has been responsible for the development, implementation and strategic management of information systems within major blue-chip organisations such as BT, Amoco and Forte. Most recently, he was the IT Manager for Gardner Merchant Leisure, a division of the world's largest catering company, Gardner Merchant Sodexho. He was until recently Vice Chair of Bradford Community Housing Trust.